\documentclass[review]{elsarticle}
\newcommand{\cmark}{\ding{51}}%
\newcommand{\xmark}{\ding{55}}%
\usepackage{lineno}
\usepackage[T1]{fontenc}
\usepackage[utf8]{inputenc}
\usepackage[bottom]{footmisc}
\usepackage{booktabs} 
\usepackage{float}
\usepackage{subfig}
\usepackage{silence}
\usepackage{color}
\usepackage{tabularx}
\usepackage{etoolbox}
\usepackage{MnSymbol}
\usepackage{hyperref}
\usepackage{amsmath}
\usepackage{listings}
\usepackage{fancybox}
\usepackage{wrapfig}
\usepackage{graphicx}
\usepackage{url}
\usepackage{paralist}
\usepackage{multirow}
\usepackage{balance}  

\newcommand{\question}[1]{\emph{\qq{#1}}}
\newcommand{\qq}[1]{``#1''}

\usepackage[normalem]{ulem}
\usepackage{lipsum}
\usepackage[flushleft]{threeparttable}
\usepackage{todonotes}

\usepackage{caption}
\usepackage{graphicx}

\usepackage{enumitem}
\setlist[enumerate]{itemsep=0mm}

\newcommand{\mycodefont}{
\fontsize{8}{9}\selectfont\ttfamily
}

\newcommand{\code}[1]{\texttt{#1}}
\modulolinenumbers[5]

\begin{document}

\begin{frontmatter}

\title{Towards Optimisation of Collaborative Question Answering over Knowledge Graphs}


\author{Kuldeep Singh*\footnote{* first two authors contributed equally}}
\ead{kuldeep.singh1@nuance.com}
\address{Nuance Communications Deutschland GmbH, Germany}

\author{Mohamad Yaser Jaradeh*}
\ead{yaser.jaradeh@tib.eu}
\address{L3S Research Center, Leibniz University of Hannover, Germany}

\author{Saeedeh Shekarpour}
\ead{sshekarpour1@udayton.edu}
\address{University of Dayton, USA}

\author{Akash Kulkarni}
\ead{akashkulkarni1192@gmail.com}
\address{Microsoft Corporation, USA}

\author{Arun Sethupat Radhakrishna}
\ead{sethu021@umn.edu}
\address{University of Minnesota, USA}

\author{Ioanna Lytra\corref{correspondingauthor}}
\ead{ioanna.lytra@cs.uni-bonn.de}
\address{University of Bonn \& Fraunhofer IAIS, Germany}

\author{Maria-Esther Vidal}
\ead{maria.vidal@tib.eu}
\address{TIB Leibniz Information Centre for Science and Technology, Germany}

\author{Jens Lehmann}
\ead{jens.lehmann@cs.uni-bonn.de}
\address{University of Bonn \& Fraunhofer IAIS, Germany}

\begin{abstract}

Collaborative Question Answering (CQA) frameworks for knowledge graphs aim at integrating existing question answering (QA) components for implementing sequences of QA tasks (i.e.\ QA pipelines). The research community has paid substantial attention to CQAs since they support reusability and scalability of the available components in addition to the flexibility of pipelines. CQA frameworks attempt to build such pipelines automatically by solving two optimisation problems: 1) local collective performance of QA components per QA task and 2)
global performance of QA pipelines. In spite offering several advantages over monolithic QA systems, the effectiveness and efficiency of CQA frameworks in answering questions is limited. In this paper, we tackle the problem of local optimisation of CQA frameworks and propose a three fold approach, which applies feature selection techniques with supervised machine learning approaches in order to identify the best performing components efficiently. We have empirically evaluated our approach over existing benchmarks and compared to existing automatic CQA frameworks. The observed results provide evidence that our approach answers a higher number of questions than the state of the art while reducing: i) the number of used features by 50\% and ii) the number of components used by 76\%.
\end{abstract}

\begin{keyword}
Question Answering, Knowledge Graph, Entity Linking, Relation Linking, Semantic Search, Experiment and Analysis
\end{keyword}

\end{frontmatter}


%

\section{Introduction}
Question Answering (QA) systems allow end users to extract useful information from several sources including documents, knowledge graphs, relational tables, etc. by posing questions in natural language or as voice input. The problem of question answering over knowledge graphs has received significant attention by the research community~\cite{GERBILQA,SHIN2019445,DBLP:journals/isci/ZhengCYZZ19,DBLP:journals/isci/HuDYY18,DBLP:conf/acl/YuYHSXZ17} since the inception of publicly available knowledge graphs such as DBpedia~\cite{swj_dbpedia}, Freebase~\cite{DBLP:conf/aaai/BollackerCT07} and Wikidata~\cite{DBLP:conf/www/Vrandecic12}. Often, a QA system over structured data acts as a black box and translates a natural language question into a formal query (e.g.\ SQL or SPARQL\footnote{\url{https://www.w3.org/TR/rdf-sparql-query/}}) to extract information from the underlying structured knowledge source. 

In the case of knowledge graphs, the formal query language is usually SPARQL. For the exemplary question "What is the timezone of India?", a QA system needs to implement several tasks such as named entity recognition and disambiguation (to map \texttt{India} to \texttt{dbr:India}\footnote{Prefix \code{dbr} is bound to \code{http://dbpedia.org/resource/}}), relation linking (e.g.\ mapping \texttt{time zone} to \texttt{dbo:timeZone}\footnote{Prefix \code{dbo} is bound to \code{http://dbpedia.org/ontology/}}) and query building to construct the corresponding SPARQL query (i.e.\ \texttt{SELECT ?c \{dbr:India dbo:timeZone ?c.\}}). Researchers have broadly adapted three approaches for building QA systems over knowledge graphs~\cite{kuldeepthesis}: 
\begin{enumerate}
  \item \textit{Semantic Parsing based QA systems}: In this approach, QA developers implement a monolithic QA system including tightly coupled modules which focus on individual stages of QA pipelines (e.g.\ named entity disambiguation, relation linking)~\cite{kuldeepthesis}. Researchers utilise the grammatical and semantic relationships between the words of the sentences and try to map those relationships to the knowledge graph concepts. Over 30 QA systems implementing such approaches, which were very popular in the last decade, have been developed~\cite{kuldeepthesis}. However, the semantic parsing based QA systems suffer from several challenges such as complex pipelines, error propagation and slower run time. 
  \item \textit{End-to-End QA systems}: With the recent advancement of machine learning technologies and growing availability of larger datasets, developers shifted focus towards proposing end-to-end neural network based QA approaches~\cite{DBLP:conf/acl/YuYHSXZ17}. These approaches skip the complex pipeline structure and focus on end-to-end mapping of natural language concepts (entities and relations) directly to knowledge graph concepts to find an answer. Most of the end-to-end QA approaches are limited to simple questions i.e.\ questions with a single entity and relation~\cite{DBLP:conf/acl/YuYHSXZ17}.
  \item \textit{Collaborative Question Answering (CQA) Frameworks}: Despite several overlapping QA tasks (e.g.\ entity linking, relation linking, etc.), reusability of QA systems for further research is limited and remains an open challenge because of the focus on specific technologies, applications or datasets. As a result, creating new QA systems is currently still cumbersome and inefficient and needs to start from scratch. CQA frameworks address this research gap and promote creating QA systems by reusing existing QA components performing various QA tasks~\cite{openqa,kimokbqa}. These frameworks follow a loosely coupled approach at the implementation level for reusing QA components for tasks such as named entity recognition and disambiguation, relation linking, etc. Therefore, existing CQA frameworks tackle scalability of QA components and allow building QA systems by arranging components performing successive QA tasks (referred as a QA pipeline). CQA frameworks often resort to semantic web technologies for integrating existing QA components to compose QA pipelines~\cite{kim2017okbqaa,DBLP:conf/coling/KimCKKC16,DBLP:conf/sigir/SinghLRVV18}; QA components can be selected either manually (e.g.\ OKBQA~\cite{DBLP:conf/coling/KimCKKC16}, openQA~\cite{openqa}) or automatically (e.g.\ Frankenstein~\cite{DBLP:conf/www/SinghRBSLUVKP0V18}). In the static CQA frameworks, a user need to manually select sequence of components (i.e. pipeline) to get the final answer of the question. The automatic CQA framework improves static CQA frameworks based on following observations 1) the performance of the QA systems and the components vary a lot based on the type of questions. For instance, on the QALD-6 dataset, CANALI QA system is the overall winner whereas when the question starts with "give me", another QA system UTQA is the winner \cite{qaldgen2019} on the subset of the dataset. 2) it is evident in the literature that question features such as question length, POS tags, question headword, etc impact the performance of a QA system \cite{loni2011survey}. 
  
An automatic CQA framework uses supervised machine learning algorithms to predict the best performing component per task for each input question. The automatic CQA frameworks such as Frankenstein \cite{DBLP:conf/www/SinghRBSLUVKP0V18} creates a labelled representation of input question using question features and trains several classifiers to predict best component per task for each input question. Thus, the label set of the training datasets for a given component was set up by measuring the micro F-Score (F-Score) of every given question. The variety of available CQA frameworks has encouraged researchers to develop high quality QA components such as EARL~\cite{DBLP:conf/semweb/DubeyBCL18}, SQG~\cite{DBLP:conf/esws/ZafarNL18} and Falcon~\cite{falcon}, focusing on improving the performance of individual QA tasks (e.g.\ entity linking, query generation) rather than building entire pipelines. Several workshops and tutorials have also been organised at different research venues with a focus on collaborative QA development\footnote{see \url{http://coling2016.okbqa.org/} and \url{http://qatutorial.sda.tech/}}~\cite{Choi:2017:SWO:3077136.3084372}.
\end{enumerate}
\subsection{Research Objective}
CQA frameworks have seen a rising interest in research and practice over the past years. Yet, the performance of the state of the art CQA framework compared to monolithic end to end QA system is limited. For instance, the baseline over QALD-5 dataset is with 0.63 Global F-score whereas state of the art CQA framework reports F-score 0.14 on the same dataset  \cite{DBLP:conf/clef/UngerFLNCCW15,DBLP:conf/www/SinghRBSLUVKP0V18}. 

This observation motivates us to tackle the problem of improving the state of the art of CQA frameworks, analyse the issues that hinder high performance of QA systems and propose corresponding solution strategies. Automatic CQA frameworks such as Frankenstein solve the global optimisation problem of finding the best performing sequence of QA components as well as the local optimisation problem of finding the best QA component for a particular task (e.g.\ named entity disambiguation). Automatic CQA frameworks rely on machine learning methods and search meta-heuristics to solve the optimisation problem of identifying the best sequence of QA components for each input question. It allows the CQA framework to select a dynamic pipeline of components for input question based on the strength of the components in answering a particular type of questions (questions containing single entity, single relation or questions with multiple entities, etc.).
Albeit overall effective in combining QA components, these frameworks perform inefficiently in terms of execution time and overall performance metric of Precision and Recall during the process of identifying the most suitable components for a QA task \cite{DBLP:conf/www/SinghRBSLUVKP0V18}. 
In this article, in order to address the inefficiency of CQA frameworks, we tackle the \textbf{local optimisation problem} and explore machine learning methods to improve the state of the art in solving this problem effectively and efficiently.   
\subsection{Approach}
We propose a three fold approach that relies on feature engineering of input questions, high-performance QA components and machine learning models (e.g.\ Random Forest, Gradient Boosting and feed-forward Neural Networks) for predicting the best performing QA component per task for each input question. We implemented our approach within Frankenstein framework to analyse its effectiveness. We name the extension of Frankenstein as Frankenstein 2.0, which is able to estimate the task performance of a QA component based on the \textit{most significant features} of an input question. The results of an extensive empirical evaluation over existing QA benchmarks indicate that the Frankenstein 2.0 prediction model not only enables the identification of the best performing QA components while using less input query features but also empowers Frankenstein to more accurately predict the best QA components per QA task.  
\subsection{Contributions}
In this article, our contributions are threefold. Firstly, we develop a feature engineering based approach that determines the most significant question features required per QA task. Secondly, we devise a prediction model based on benchmarking of supervised learning models, which is able to exploit the selected features per task to find the best performing components for an input question. Thirdly, we also integrate recently released high performing QA components implementing various QA tasks in Frankenstein. We report on the results of an extensive empirical study showing the overall impact of our approach. The observed results suggest that our approach implemented as Frankenstein 2.0 outperforms the previous version of the framework in terms of efficiency and effectiveness.

The article is organised as follows: Section~\ref{related} summarises the related work. Section~\ref{sec:frank} details the background about Frankenstein framework. Section~\ref{problem} describes the problem tackled in the scope of the current paper and presents the proposed approach. Section~\ref{experiments} includes detailed experiments for the evaluation of the proposed approach. Finally, we discuss conclusions in Section~\ref{sec:Conclusions}.

\section{Related Work} \label{related}
Question answering over knowledge graphs has gained momentum in the last decade and researchers from different communities, e.g.\ semantic web, information science, databases and natural language processing have extensively studied this problem and proposed several QA systems~\cite{GERBILQA,DBLP:journals/isci/ZhengCYZZ19,DBLP:journals/isci/HuDYY18,DBLP:conf/www/UngerBLNGC12,DBLP:journals/isci/FerrandezTFFM09}. 
DBpedia is the prominent background knowledge graph in this setting, and researchers have developed more than 35 QA systems over DBpedia (detailed in~\cite{GERBILQA} and~\cite{kuldeepthesis}). 
Although these QA systems achieved high performance on specific data sets, they expose limitations in reusability due to their monolithic implementations. To promote reusability within the QA community, researchers have attempted to build modular frameworks to allow researchers to improve individual stages of a QA pipeline and reused components with higher accuracy for other QA tasks. QALL-ME~\cite{qallme} is one of the initial attempts in this direction that provides a platform for building multilingual QA systems. The openQA framework~\cite{openqa} combines several QA components and existing QA systems like SINA~\cite{DBLP:journals/ws/ShekarpourMNA15} in its architecture. The main downside of openQA is a strict programming language requirement. The OKBQA framework~\cite{DBLP:conf/coling/KimCKKC16} overcomes this problem and provides a RESTful API based architecture to build QA systems; it has 24 components implementing various QA tasks such as template generation, disambiguation, query building and answer generation. These three are manual CQA frameworks, i.e.\ users have to select the components to execute QA pipelines. With an increasing number of QA components developed by the research community, manual QA frameworks do not address the scalability of components for various QA tasks. With the availability of many QA components in a single platform, it is not clear if it is expected to run all the possible viable combinations for each input question. For example, if a QA framework has X number of components for named entity disambiguation (NED), Y for relation linking (RL), Z for query builder (QB) task, the number of resulting pipelines in the framework is XYZ ($X*Y*Z$). In existing manual frameworks, there is no dynamic (on the fly) way to select the best components per task to be part of the QA pipeline for the given input question based on the strengths and weaknesses of these components.

The recently released Frankenstein framework~\cite{DBLP:conf/www/SinghRBSLUVKP0V18,DBLP:conf/sigir/SinghLRVV18} provides a domain agnostic platform for developers to build QA pipelines in a collaborative manner (for domain agnostic feature, please see~\cite{geoqa}). It is built using a formal methodology called Qanary~\cite{singhqanary,DBLP:conf/esws/SinghBDSC016} which utilises an RDF\footnote{\url{https://www.w3.org/RDF/}} vocabulary (QA vocabulary~\cite{DBLP:conf/semco/SinghBDS16}) to provide homogeneous data exchange between components. It is an automatic CQA framework that implements the optimisation problem of selecting the best component per task to create effective QA pipelines. Frankenstein implements logistic regression classifiers for selecting best-performing components for a given question based on \textit{all the features of a given input question and all the available components}. Therefore, Frankenstein blindly traverses the space of potential components for a task and is not able to differentiate the impact of a question feature such as question length, POS tags, question head word etc. Since the number of features can be large, considering all the features together may have a negative impact on the performance of CQA frameworks. Our proposed three fold approach implemented in Frankenstein 2.0 framework also provides an automatic way to select the best performing QA components per task to implement a QA pipeline; however, contrary to Frankenstein, Frankenstein 2.0 is able to select the most significant feature for a given question. In some cases, reduction of features can impact up to 50\% positively on the overall performance.

\section{Frankenstein Framework} 
\label{sec:frank}
\begin{table}[tb!]
	\centering
	\resizebox{0.99\columnwidth}{!}{
       \begin{tabular}{ l l l l l }
   	    \toprule
            \textbf{Functionality} & \textbf{Frankenstein} & \textbf{QALL-ME} & \textbf{openQA} & \textbf{OKBQA} \\
            \midrule
            {\it Promotes Reusability}
               & $\checkmark$ & $\checkmark$ & $\checkmark$ & $\checkmark$ \\
            {\it Programming Language Independent}
               & $\checkmark$ & - & - & $\checkmark$ \\
           {\it Input/Output Format Independent}
               & $\checkmark$ & - & - & - \\
           {\it Number of Reusable Components}
                & 29 & 7 & 2 & 24 \\
            {\it Automatic QA pipeline Composition}
                & $\checkmark$ & - & - & - \\
            {\it Microservice Based Architecture}
                & $\checkmark$ & $\checkmark$ & - & $\checkmark$ \\
            {\it Use of Semantic Web Technologies}
                & $\checkmark$ & $\checkmark$ & $\checkmark$ & $\checkmark$ \\
            \bottomrule
        \end{tabular}
        }
        \caption{Comparison of Various CQA Frameworks}
    \label{tab:qaframeworkss}
\end{table} 

Frankenstein is the first framework of its kind for integrating all state-of-the-art QA components to build more powerful QA systems with collaborative efforts. The comparison of various functionalities of Frankenstein with other QA frameworks including QALL-ME framework~\cite{qallme}, OKBQA~\cite{kimokbqa} and openQA~\cite{DBLP:conf/semweb/MarxSENL15} is given in Table~\ref{tab:qaframeworkss}. Unlike other CQA frameworks, Frankenstein simplifies the integration of emerging components and is sensitive to the input question.
The rationale is instead of building a QA system from scratch to rather reuse currently existing QA components available to the QA community. Each heterogeneous component is integrated as micro service and for every input question, a local knowledge graph using QA vocabulary~\cite{DBLP:conf/semco/SinghBDS16} is created to store all the information in a knowledge base with a unique graph ID. This knowledge graph stores information such as question, annotation of various parts of input question (e.g.\ entity, relation, class and provenance information) which can be used to analyse the output of individual stages of a QA pipeline. Hence, Frankenstein introduces a concept of \textbf{knowledge driven service oriented architecture}. Frankenstein not only integrate several components in the architecture, but also proposes a novel solution to choose a pipeline consisting of best components per task to answer an input question. Frankenstein supersedes other QA frameworks in integrating a number of components and various offered functionalities as illustrated in Table~\ref{tab:qaframeworkss}. It trains several classifiers based on question features as label set (e.g.\ question length, POS tags, answer type) against F-score of each component per question in order to predict the performance of the component for the input question. It builds a dynamic pipeline per question using some of the components from the pool of available components. Frankenstein is domain agnostic, and researchers have reused Frankenstein to build geospatial question answering systems by adding new components for specific geospatial functionalities and reusing NED components from existing implementations~\cite{geoqa}. Therefore, Frankenstein provides a smart solution to build QA systems collaboratively.

\subsection{Prediction of Best Component}
In this context, we formalise set of necessary QA tasks as $\mathcal{T}=\{t_1,t_2,\dots,t_n\}$ such as entity disambiguation, relation linking, etc.
Each task ($t_i:q^* \rightarrow q^+ $) transforms a given representation $q^*$ of a question $q$ into another representation $q^+$.
For example, NED and RL tasks transform the input representation \question{What is the timezone of India?}\ into the representation \question{What is the \code{dbo:timeZone} of \code{dbr:India}?}.

The performance of an automatic CQA pipeline depends on two optimisation tasks which have been formally defined for Frankenstein~\cite{DBLP:conf/www/SinghRBSLUVKP0V18}:
\begin{enumerate}
    \item \textbf{Local optimisation:} the problem of finding the best performing component for accomplishing the task $t_i$ for an input question $q$, denoted as $\gamma^{t_i}_{q}$, is formulated as follows:
\begin{equation}\label{eq:eque3}
\gamma^{t_i}_{q} = \arg\max_{C_j \in \mathcal{C}^{t_i}} \{\mathit{Pr}(\rho(C_j)|q) \}
\end{equation}

Where $\mathit{Pr}(\rho(C_j)|q,t_i)$ is a supervised learning problem to predict the performance of the given component $C_j$ for the given question $q$; 
Please note that the entire set of 
QA components is denoted by $\mathcal{C}=\{C_1,C_2,\dots,C_m\}$. 
Each component $C_j$ solves one single QA task; $\theta(C_j)$ corresponds to the QA task $t_i$ in $\mathcal{T}$ implemented by $C_j$.
For example, Tagme implements the NED QA task, i.e.  $\theta(\mathit{Tagme})=\mathit{NED}$. 

\paragraph{Solution} Suppose we are given a set of questions $\mathcal{Q}$ with the detailed results of performance for each component per task.
We can then model the prediction goal $\mathit{Pr}(\rho(C_j)|q,t_i)$ as a supervised learning problem on a training set,  
i.e.  a set of questions $\mathcal{Q}$ and a set of labels $\mathcal{L}$ representing the performance of $C_j$ for a question $q$ and a task $t_i$. 
In other words, for each individual task $t_i$ and component $C_j$, the purpose is to train a supervised model that predicts the performance of the given component $C_j$ for a given question $q$ and task $t_i$ leveraging the training set.
If $|\mathcal{T}|=n$ and each task is performed by $m$ components, then $n \times m$ individual learning models have to be built up.
Furthermore, since the input questions $q \in \mathcal{Q}$ have a textual representation, it is necessary to automatically extract suitable features from the question, i.e.  $\mathcal{F}(q)=(f_1,\dots,f_r)$.

\item \textbf{Global optimisation:} the problem of finding the best performing pipeline of QA components $\psi_q^\mathit{goal}$, for a  question $q$ and a set of QA tasks called $\mathit{goal}$. Formally, this optimisation problem is defined as follows:
\begin{equation}\label{eq:eque4}
\psi_q^\mathit{goal} = \arg\max_{ \eta \in \mathcal{E}(goal)} \{\Omega(\eta,q)\}
\end{equation}
where $\mathcal{E}(\mathit{goal})$ represents the set of pipelines of QA components that implement $\mathit{goal}$ and $\Omega(\eta,q)$ corresponds to the estimated performance of the pipeline $\eta$ on the question $q$.
\end{enumerate}

\paragraph{Solution} Frankenstein proposes a greedy algorithm that relies on the \textit{optimisation principle} that states that an optimal pipeline for a goal and a question $q$ is composed of the best performing components that implement the tasks of the goal for $q$.
Suppose that $\oplus$ denotes the composition of QA components, then an 
optimal pipeline $\psi_q^{goal}$ is defined as follows:
\begin{equation}\label{eq:eque5}
 \psi_q^\mathit{goal} := \oplus_{t_i\in\mathit{goal}} \{\gamma_q^{t_i}\} 
\end{equation}

The proposed greedy algorithm works in two steps: \textit{QA Component Selection} and \textit{QA Pipeline Generation}. During the first step of the algorithm, each task $t_i$ in $\mathit{goal}$ is considered in isolation to determine the best performing QA components that implement $t_i$ for $q$, i.e.  $\gamma_q^{t_i}$. 
For each $t_i$ an ordered set of QA components is created based on the performance predicted by the supervised models that learned to solve the problem described in \autoref{eq:eque3}. 
Consider the question $q$=\question{What is the timezone of India?} and $\mathit{goal}=\{\mathit{NED},\mathit{RL},\mathit{QB}\}$. 
The algorithm aims to create an ordered set $\mathit{OS}_{t_{i}}$ of QA components for each task $t_i$ in $\mathit{goal}$. 
Components are ordered in each $\mathit{OS}_{t_{i}}$ according to the values of the performance function $\rho(.)$ predicted by the supervised method trained for questions with the features $\mathcal{F}(q)$ and task $t_i$; in our example,  $\mathcal{F}(q)$=\{(\text{QuestionType}:\text{What}),
(\text{AnswerType}:\text{String}), (\text{\#words}:6), (\text{\#DT}:1), (\text{\#IN}:1), (\text{\#WP}:1), (\text{\#VBZ}:1), (\text{\#NNP}:1), (\text{\#NN}:1)\} indicates that $q$ is a \textit{WHAT} question whose answer is a \textit{String}; further, $q$ has six words and POS tags such as determiner, noun etc.
Based on this information, the algorithm creates three ordered sets:  $\mathit{OS}_\mathit{NED}$, $\mathit{OS}_\mathit{RL}$, and $\mathit{OS}_\mathit{QB}$. 
 
In the second step, the algorithm follows the optimisation principle in \autoref{eq:eque5} and combines the top $k_i$ best performing QA components of each ordered set. 

\subsection{Frankenstein Architecture}
The following modules are part of the Frankenstein architecture:
~\\
{\bf Feature Extractor}. This module derives a set of features from an input question.
Features include, for instance, question length, question and answer types and POS tags.
~\\
{\bf QA Components}. Frankenstein in the original implementation integrates 29 QA components implementing five QA tasks, namely Named Entity Recognition (NER), Named Entity Disambiguation (NED), Relation Linking (RL), Class Linking (CL) and Query Building (QB). In most of the questions NED, RL and QB components are necessary to generate the SPARQL query for the input question. Sometimes, to formulate a SPARQL query for a given question, it is necessary to also disambiguate classes against the ontology.
For example, in the question \question{Which river flows through Seoul}, \question{river} needs to be mapped to \texttt{dbo:River}\footnote{\url{http://dbpedia.org/ontology/River}}.
Table~\ref{tab:components} provides a list of QA components integrated in Frankenstein. The 11 NER components are used with AGDISTIS~\cite{UsbeckNRGCAB14} to disambiguate entities as AGDISTIS requires the question and spotted position of entities as input. Henceforth, any reference to NER tool, will refer to its combination with AGDISTIS, and we have excluded individual performance analysis of NER components. However, other 7  NED components recognise and disambiguate the entities directly from the input question. Hence, Frankenstein has 18 NED, 5 RL, 2 CL, 2 QB components. 
~\\
{\bf QA Component Classifiers}. For each QA component, an independent Classifier is trained; it learns from a set of features of a question and predicts the performance of a particular component.
~\\
{\bf QA Pipeline optimiser}. Pipeline optimisation is performed by two modules.
The {\bf Component Selector} uses the best performing components for accomplishing a given task based on the input features and the results of the QA Component Classifiers; the selected QA components are afterwards passed to the {\bf Pipeline Generator} to automatically generate the corresponding QA pipelines.
~\\
{\bf Pipeline Executor}. This module is used to extract answers from the underlying knowledge graph (DBpedia in this case) using the best predicted pipeline.

\begin{table}[htb!]
	\centering
	\resizebox{.8\textwidth}{!}{%
	  \begin{threeparttable}
		\begin{tabular}{ l l l l l l }
			\toprule
			\textbf{Component/} & \textbf{QA Task} & \textbf{Year} & \textbf{Open} & \textbf{RESTful} & \textbf{Publi-} \\
			\textbf{Tool} & & & \textbf{Source} & \textbf{Service} & \textbf{cation} \\
			\midrule
			\textit{Entity Classifier}~\cite{Dojchinovski:2013:ECMLPKDD13}
			& NER & 2013 & \xmark & \cmark & \cmark \\
			
			\textit{Stanford NLP}~\cite{DBLP:conf/acl/FinkelGM05}
			& NER & 2005 & \cmark & \cmark & \cmark \\
			
			\textit{Ambiverse}~\cite{DBLP:conf/lrec/RizzoET14}\tnote{i}
			& NER/NED & 2014 & \xmark & \cmark & \cmark \\
			
			\textit{Babelfy}~\cite{DBLP:journals/tacl/0001RN14}\tnote{ii}
			& NER/NED & 2014 & \xmark & \cmark & \cmark \\
			
			\textit{AGDISTIS}~\cite{UsbeckNRGCAB14}
			& NED & 2014 & \cmark & \cmark & \cmark \\
			
			\textit{MeaningCloud}~\cite{DBLP:journals/cii/MartinezMSSLR16}\tnote{iii}
			& NER/NED & 2016 & \xmark & \cmark & \cmark \\
			
			\textit{DBpedia Spotlight}~\cite{MendesJGB11}
			& NER/NED & 2011 & \cmark & \cmark & \cmark \\
			
			\textit{Tag Me API}~\cite{DBLP:journals/software/FerraginaS12}
			& NER/NED & 2012 & \cmark & \cmark & \cmark \\
			
			\textit{Aylien API}\tnote{iv}
			& NER/NED & - & \xmark & \cmark & \xmark \\
			
			\textit{TextRazor}\tnote{v}
			& NER  & - & \xmark & \cmark & \xmark \\
			
			\textit{OntoText}~\cite{simov2016role}\tnote{vi}
			& NER/NED & - & \xmark & \cmark & \cmark \\
			
			\textit{Dandelion}\tnote{vii}
			& NER/NED & - & \xmark & \cmark & \xmark \\
			
			\textit{RelationMatcher}~\cite{kcap}
			& RL & 2017 & \cmark & \cmark & \cmark \\
			
			\textit{ReMatch}~\cite{DBLP:conf/i-semantics/MulangSO17}
			& RL & 2017 & \cmark & \cmark & \cmark \\
			
			\textit{RelMatch}~\cite{kimokbqa}
			& RL & 2017 & \cmark & \cmark & \cmark \\
			
			\textit{RNLIWOD}\tnote{viii}
			& RL & 2016 & \cmark & \xmark & \xmark \\
			
			\textit{Spot Property}~\cite{kimokbqa}\tnote{ix}
			& RL & 2017 & \cmark & \cmark & \cmark \\
			
			\textit{OKBQA DM CLS}\tnote{ix}
			& CL & 2017 & \cmark & \cmark & \cmark \\
			
			\textit{NLIWOD CLS}\tnote{viii}
			& CL & 2016 & \cmark & \xmark & \xmark \\
			
			\textit{SINA}~\cite{DBLP:journals/ws/ShekarpourMNA15}
			& QB & 2013 & \cmark & \xmark & \cmark \\
			
			\textit{NLIWOD QB}\tnote{viii}
			& QB & 2016 & \cmark & \xmark & \xmark \\
			\bottomrule
		\end{tabular}
		\begin{tablenotes}
            \item[i] \url{https://developer.ambiverse.com/}
            \item[ii] \url{https://github.com/dbpedia-spotlight/dbpedia-spotlight}
            \item[iii] \url{https://www.meaningcloud.com/developer}
            \item[iv] \url{http://docs.aylien.com/docs/introduction}
            \item[v] \url{https://www.textrazor.com/docs/rest}
            \item[vi] \url{http://docs.s4.ontotext.com/display/S4docs/REST+APIs}
            \item[vii] \url{https://dandelion.eu/docs/api/datatxt/nex/getting-started/}
            \item[viii] \url{https://github.com/dice-group/NLIWOD}
            \item[ix] \url{http://repository.okbqa.org/components/7}
        \end{tablenotes}
	  \end{threeparttable}
	}
		\caption{{\bf Components Integrated in Frankenstein}: 8 QA components are not available as open source software, 24 provide a RESTful service API and 22 are accompanied by peer-reviewed publications.}
	\label{tab:components}    
\end{table}
\section{Problem Statement and Approach} \label{problem}

In this paper, we focus \textbf{only} on the local optimisation of the CQA frameworks (Cf. \autoref{eq:eque3}). Our local optimisation approach relies on the performance of a given QA component (denoted by $\rho(C_j)$) and a prediction approach which estimates the performance of components (denoted as $\mathit{Pr}(\rho(C_j)|q)$).
Thus, to achieve the optimum local optimisation, we apply three categories of enhancements: (i) study the impact of question features on the performance and develop a feature selection module, (ii) build up learning models with high performance per component and (iii) integrate components with high F-score.
In the following subsections, we discuss the associated challenges followed by proposed solution strategies.

\subsection{Reliance of on Learning Models}
Frankenstein predicts best performing QA components per task using individual classifiers trained per component to generate QA pipelines. 
Thus, we initially investigate the bottlenecks of these classifiers. 
There are three typical bottlenecks that influence the performance of a given classifier, namely, (i) the quality of training data set, (ii) the feature set and (iii) the learning model.
In the following paragraphs, we discuss these bottlenecks in more detail.

\textbf{(i) Quality of Training Data Set:} 
To have a fair judgement, it is required that the benchmark data set contains diverse and relatively even types of questions (e.g.\ simple versus complicated questions and short versus long questions). 
The other concern is related to the number of positive samples versus negative samples taken into account in the training data sets for training the classifiers.
For every given component, all the questions answered are considered as positive samples and the rest as the negative samples. This ratio is skewed since the majority of components demonstrate poor performance (i.e.\ number of negative samples is far higher)~\cite{DBLP:journals/corr/abs-1809-10044}.

\emph{Strategy:} 
The first step of our three fold approach is to enhance the quality of the underlying benchmark data set by (i) balancing questions from diverse types and (ii) balancing the number of positive samples versus negative samples.

\textbf{(ii) Feature Set}
The classifiers built up in the Frankenstein framework rely on abstract and primary features. 
This limited feature set might not be sufficient to represent all the semantics and structure of an input question. 

\emph{Strategy (feature engineering):}
The second step of our three fold approach is to develop a feature engineering technique that includes new features such as word embeddings which have recently demonstrated higher quality with regard to the proper encoding of the structure and semantic patterns~\cite{word2vec}. 
Further, we demonstrated the positive impact of feature engineering approach by implementing a feature selection module in Frankenstein 2.0. Frankenstein 2.0 runs feature engineering approaches to find out a reduced number of features which leads to better performance.
In particular, Frankenstein 2.0 uses two methods for feature selection, Recursive Feature Elimination (RFE) and Extremely Randomised Trees (ERT).

\textbf{(iii) Learning Models:}
Naturally, the selection of the learning model will influence the overall performance of the approach. 
\\
\emph{Strategy:} 
The third step of our approach is to provide a benchmarking approach across several supervised learning models to find out the best-performing model. 

\subsection{Dependency to External Components}
\textbf{Challenge 1:} The performance of the integrated components in CQA frameworks $\rho(C_i)$ plays a major role in the local optimisation.
If CQA frameworks integrate components with low performance, even in case of applying the best prediction and optimisation algorithms, the output performance will be poor. In Frankenstein this issue causes limited overall performance of the framework~\cite{DBLP:conf/www/SinghRBSLUVKP0V18}.

\textbf{Strategy: Integrating new components.} The research community has started working in the direction of creating high-performance components for the different QA tasks in order to address question answering with collaborative efforts. Only during last year, several QA components have been released explicitly for CQA frameworks such as Falcon~\cite{falcon}\footnote{\url{https://labs.tib.eu/falcon/}}, EARL~\cite{DBLP:conf/semweb/DubeyBCL18} etc. However, existing query builder and class linking components are still poor in terms of runtime and F-score, which hinders the overall global improvement (at complete pipeline level) of CQA frameworks~\cite{DBLP:journals/corr/abs-1809-10044,DBLP:conf/coling/KimCKKC16}\footnote{the complete pipeline of OKBQA \url{http://demo.okbqa.org/} and Frankenstein is limited in resulting answers}.

\subsection{Frankenstein 2.0 Architecture}
\begin{figure*}[t]
	\centering
	\includegraphics[width=.99\textwidth]{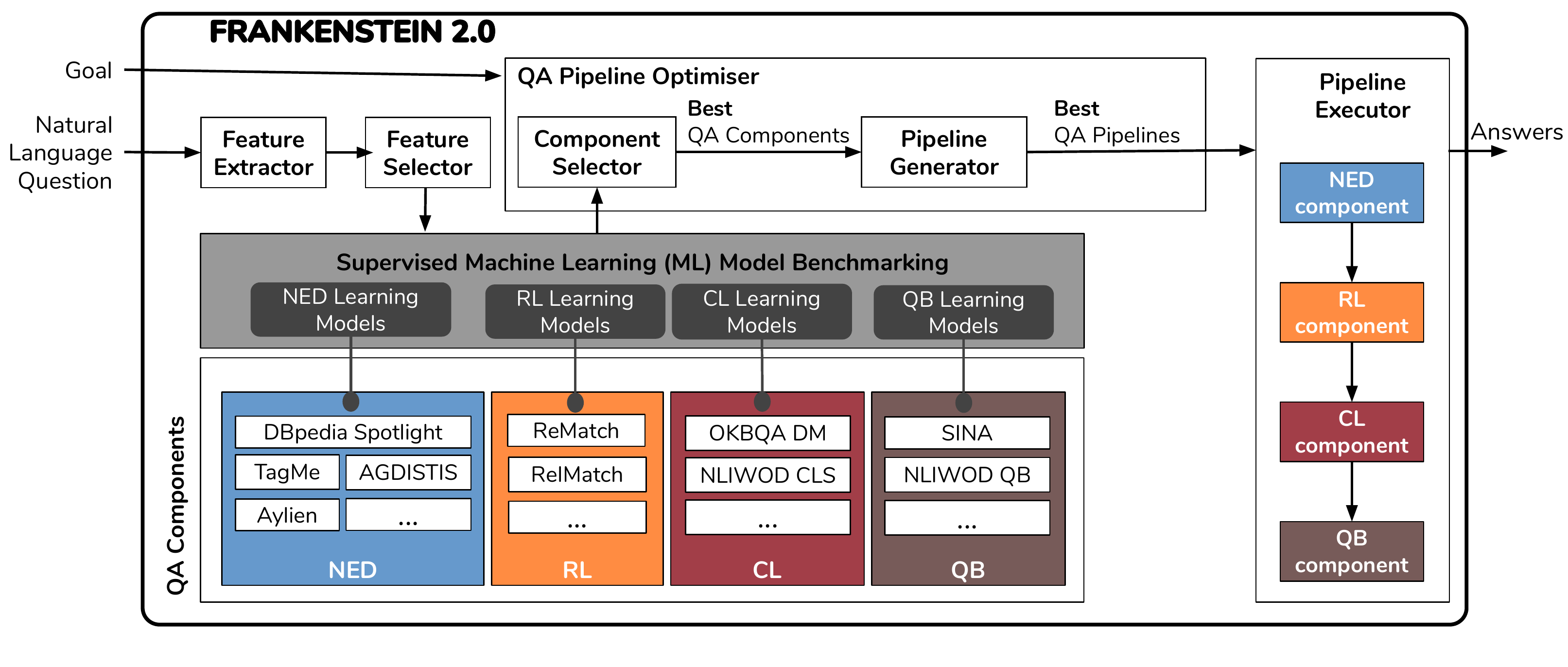}
	\caption{\textbf{Frankenstein 2.0 architecture}. The feature selection module and supervised learning model benchmarking layer have been newly added. Rest all modules belong to first version of  Frankenstein.}
	\label{fig:archi2}
\end{figure*}

 We implemented Frankenstein 2.0 extending prior implementation of Frankenstein in three directions: (i) we improved the feature extractor and selector module, (ii) we implemented a new module for applying new supervised learning models and (iii) we added newly released components such as EARL~\cite{DBLP:conf/semweb/DubeyBCL18}, Falcon~\cite{falcon} and Ambiverse~\cite{DBLP:conf/emnlp/HoffartYBFPSTTW11}. The architecture diagram of Frankenstein 2.0 is depicted in Figure \ref{fig:archi2}.
 
\section{Experimental Study} \label{experiments}
\label{sec:study}

{\bf Knowledge Graph.} \begin{inparaenum}[\itshape i\upshape)]
We employ DBpedia\footnote{DBpedia version 2016-10}~\cite{DBLP:conf/semweb/AuerBKLCI07} as underlying knowledge base which contains more than 5.6 million entities and more than 111 million RDF triples. Its size is 14.2 GB. 
\end{inparaenum}
~\\
{\bf Data sets.} \begin{inparaenum}[\itshape i\upshape)]
We rely on LC-QuAD~\cite{trivedi2017lc} data set tailored to DBpedia. LC-QuAD has 5,000 questions. However, only 3,253 questions were utilised by Frankenstein experimental study in~\cite{DBLP:conf/www/SinghRBSLUVKP0V18}. To provide a fair comparative study, we take the same questions into account. 
\end{inparaenum}
~\\
{\bf Implementation Details.} \begin{inparaenum}[\itshape i\upshape)]
We ran our experiments on a virtual server, with eight cores, 32 GB RAM running on the Ubuntu 16.04.3 operating system. We utilised the open source implementation of Frankenstein as underlying platform released in~\cite{DBLP:conf/esws/SinghBRS18}.
Frankenstein 2.0 was implemented in Python 3.6. 
Please note that for brevity, we report Frankenstein 2.0's \textbf{best setting results} in the paper. For each experiment mentioned in Table \ref{tab:10foldfeature}, Table \ref{tab:newcomponent} and Table \ref{tab:newmodel}, and the extended results for all other settings as well as the source code can be found in our public Github. We also executed our experiments on a balanced data set and overall results for each experiment were comparable but not surprisingly superior. The results in the tables are on average for all folds over 10 fold cross-validation.

We rely on the following metrics in our experiments for measuring the performance (reported by Singh et al.~\cite{DBLP:conf/www/SinghRBSLUVKP0V18} as well):
\end{inparaenum}
~\\
\begin{inparaenum}[\itshape i\upshape)]
\item \texttt{Total Questions per fold (\#totalquestions}): total number of questions in each fold while doing K-fold cross-validation.
\item \texttt{Answerable Questions\\(\#answerable}): the average number of questions in each fold (in K-fold cross-validation) for which at least one of the components per questions has an F-Score greater than $0.5$. 
\item \texttt{Predicted Top N Questions}: the average number of questions for which at least one of the predicted Top N components selected by the Classifier has an F-Score greater than $0.5$.
\item \texttt{Baseline}: Frankensteins' original setting with exactly same number of components per task (18 for Named Entity Disambiguation (NED) task, five for Relation Linking (RL) task, two Class Linking (CL) task\footnote{CL components map ontology classes \url{https://www.w3.org/2002/07/owl\#Class} present in the input question which are often required to formulate the correct SPARQL query}, two for Query Builder task (QB)), same number of features (28 total question features) and logistic regression model as prediction model.
\end{inparaenum}

\begin{figure*}[t]
	\centering
	\subfloat{\includegraphics[width=.5\textwidth]{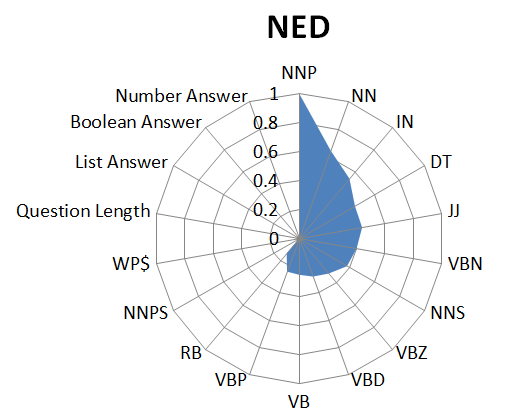}}
	\hfill
	\subfloat{\includegraphics[width=.5\textwidth]{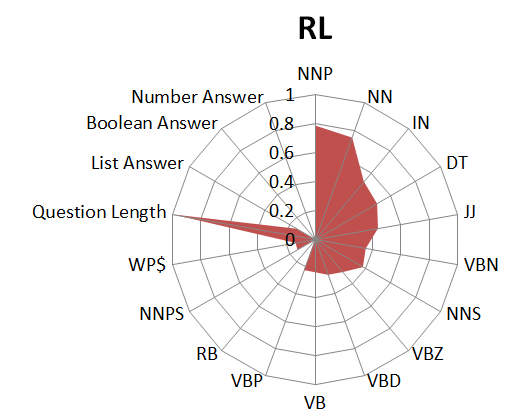}}
	\hfill
	\subfloat{\includegraphics[width=.5\textwidth]{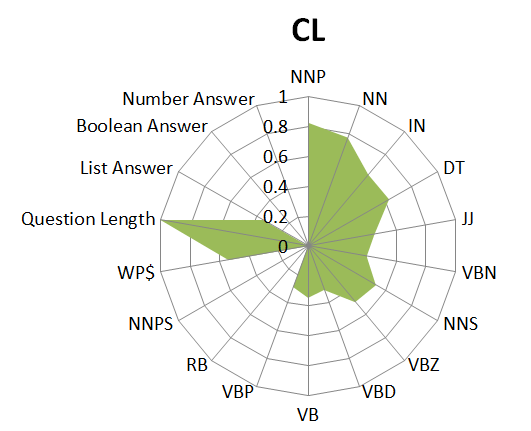}}
	\hfill
	\subfloat{\includegraphics[width=.5\textwidth]{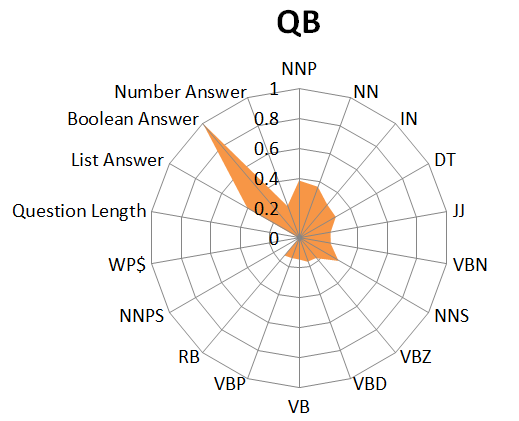}}
	\caption{
	\textbf{Significance of Features per QA task.} The figure illustrates 18 features across all tasks, e.g.\ Named Entity Disambiguation (NED), Relation Linking (RL), Class Linking (CL) and Query Builder (QB) based on the Gini importance (higher is better). A large number of features are irrelevant per task, e.g.\ for NED six out of 18 features are irrelevant.
	}
	\label{fig:old-gini}
\end{figure*}

\subsection{Experiment 1: Impact of Feature Engineering on Local Optimisation} \label{sec:experiment1}
\label{sec:feature}

In this experiment, we pursue the research question \emph{R1: what is the impact of feature engineering on local optimisation?} To study this impact, we evaluate the impact of the feature selection module of Frankenstein 2.0 on local optimisation per QA task. For the same, we did not change the machine learning module and the number of components per task of the baseline.
We separately study this impact for each task. The study presented in~\cite{DBLP:journals/corr/abs-1809-10044} showed the impact of character cases and entity type on QA component performance. Also, we represented input questions using word embeddings using state-of-the-art approaches such as Fasttext~\cite{bojanowski2017enriching}. We list the entire set of features grouped into multiple configurations:
\begin{enumerate}
    \item CF1: Basic NLP features (question length, POS-tags, Answer type) used in baseline. Total number of features are 28.
    \item CF2: Addition to CF1, we include character cases and entity type as new NLP features (entity type is excluded for NED task to avoid bias). It resulted into 51 total number of features.
    \item CF3: Phrase embedding: It is calculated by averaging the word embeddings for all words in question. 
    \item CF4: Question embedding: by concatenating all of the word embeddings of the question.
    \item CF5: Phrase vector embedding without stop-words: similar to CF3 but with the removal of stop-words (is, was, of, etc.) from the question.
    \item CF6: Combination of CF1 and CF3.
\end{enumerate}
To estimate the importance of features, we rely on a metric called Mean Decrease Gini which is a measure of variable importance for estimating a target variable and provides an indication of node impurity ~\cite{conf/kdd/TruongLB04}. For each feature, Gini importance is computed and then top-N features are selected. To select top N features, we executed RFE~\cite{journals/corr/WuGB16} and ERT~\cite{Geurts2006} feature selection techniques for each feature configuration CF1--CF6 (12 experimental settings in total considering CF1--CF6 with RFE and ERT).
For illustrating our results, we have the following three settings:
\begin{enumerate}
    \item Baseline: results of local optimisation (task level experiments) of Frankenstein~\cite{DBLP:conf/www/SinghRBSLUVKP0V18} where top N questions per task are predicted using logistic regression classifiers and features from CF1.
    \item Frankenstein 2.0$_{F}$: Baseline setting with CF2 feature configuration (i.e.\ new features per task).
    \item Frankenstein 2.0$_{FS}$: applying feature selection method on CF2 to select top N most impacting features per task. We empirically chose N=15.
\end{enumerate}


\begin{table}[tb] 
	\centering
	\caption{Feature selection impact. Effectiveness of Frankenstein 2.0 is reported based on predicted top N components per task. QB and CL just have two components while NED and RL has 18 and five, respectively. Considering most significant features of input question allows Frankenstein 2.0 to outperform the baseline. }
	\resizebox{\columnwidth}{!}{%
	    \begin{tabular}{ l c c c c c c}
	        \toprule
		    \textbf{QA} & \multicolumn{1}{l}{\textbf{\#totalquestions}} & \multicolumn{1}{l}{\textbf{\#answer-}} & \multicolumn{3}{c}{\textbf{\#Predicted}} & \multicolumn{1}{l}{\textbf{Setting}} \\
		    \cline{4-6}
		    \textbf{Task} & \multicolumn{1}{l}{} & \multicolumn{1}{l}{\textbf{able}} & \multicolumn{1}{l}{\textbf{Top1}} & \multicolumn{1}{l}{\textbf{Top2}} & \multicolumn{1}{l}{\textbf{Top3}} & \multicolumn{1}{l}{\textbf{}} \\ 
		    \midrule
		    NED & 324.3 & 294.2 & 245.2 & 270.9 & 284.3 & \multicolumn{1}{c}{Baseline}  \\
		    NED & 324.3 & 294.2 & \underline{247.1}& 270.7 & 284.3 & \multicolumn{1}{c}{Frankenstein 2.0$_{F}$}  \\
		    NED & 324.3 & 294.2 & \underline{250.4} & 271.5 & 284.3 & \multicolumn{1}{c}{\textbf{Frankenstein 2.0$_{FS}$}}  \\
		    RL & 324.3 & 153.1 & 90.3 & 118.9 & 134.4 & \multicolumn{1}{c}{Baseline}  \\
		    RL & 324.3 & 153.1 &\underline{89.1}& 118.9 & 134.4 & \multicolumn{1}{c}{Frankenstein 2.0$_{F}$}  \\
		    RL & 324.3 & 153.1 & \underline{91.4} & 120.3 & 134.6 & \multicolumn{1}{c}{\textbf{Frankenstein 2.0$_{FS}$}}  \\
		    CL & 324.3 & 76 & 68.1 & 76 & \multicolumn{1}{c}{–}& \multicolumn{1}{c}{Baseline}  \\
		    CL & 324.3 & 76 & \underline{68.4} & 76 & \multicolumn{1}{c}{–}& \multicolumn{1}{c}{Frankenstein 2.0$_{F}$}  \\
		    CL & 324.3 & 76 & \underline{68.4} & 76 & \multicolumn{1}{c}{–}& \multicolumn{1}{c}{\textbf{Frankenstein 2.0$_{FS}$}}  \\
		    QB & 324.3 & 175.4 & 162.7 & 175.4 & \multicolumn{1}{c}{–}& \multicolumn{1}{c}{Baseline} \\
		    QB & 324.3 & 175.4 & \underline{162.9}& 175.4 & \multicolumn{1}{c}{–}& \multicolumn{1}{c}{Frankenstein 2.0$_{F}$} \\
		    QB & 324.3 & 175.4 & \underline{162.9} & 175.4 & \multicolumn{1}{c}{–}& \multicolumn{1}{c}{\textbf{Frankenstein 2.0$_{FS}$}}  \\
		    \bottomrule
		\end{tabular}
		}
	\label{tab:10foldfeature}    
\end{table}

\begin{figure*}[t]
	\centering
	\includegraphics[width=.94\textwidth]{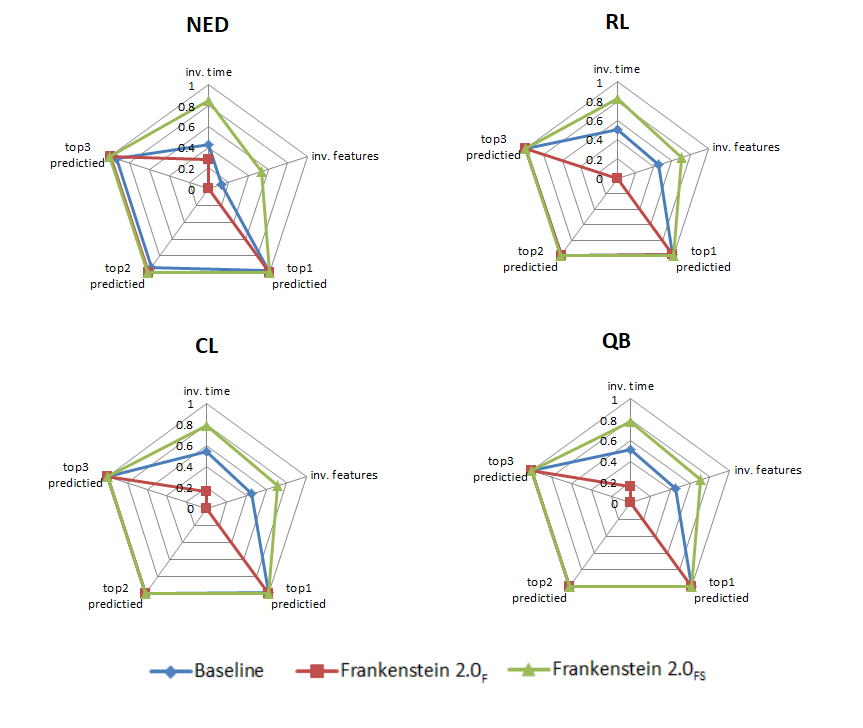}
	 \caption{\textbf{The effectiveness and efficiency of feature selection per task}. Effectiveness is measured in normalised top N predicted questions per task. While efficiency is measured based on the inverse of the normalised number of selected features and normalised execution time taken by the prediction model. For all metrics, higher is better. Inverse execution time has been increased up to 80\%, the inverse number of features is increased up to 50\%. Selecting the most significant features not only reduces execution time but also improves the number of predicted top N questions per task.}
	\label{fig:old-gini11}
\end{figure*}

 We tested with top 5, 10, 15, 20, 25 features and found that with top 15 impacting features per task, we could replicate same or slightly better results (Frankenstein 2.0$_{FS}$ in Table \ref{tab:10foldfeature}) from baseline. Table \ref{tab:10foldfeature} summarises our results in three different settings. We empirically observe that among all feature configurations (CF1-CF6), CF2 has most impact across tasks. In \figurename~\ref{fig:old-gini} which illustrates the Gini importance of features, it is clearly visible that few features have the most impact and few do not have any impact on all tasks. We compare Frankenstein 2.0$_{FS}$ setting with baseline using the following metrics: 1) execution time to train and select prediction model, 2)number of selected features and 3) top N correctly predicted questions by prediction model. Frankenstein 2.0$_{FS}$ demonstrates overall improvement as illustrated in \figurename~\ref{fig:old-gini11}. \\
 \textbf{Experiment Conclusion}: We conclude that there are different features per task that impact the performance of the learning module at local optimisation and several features do not have any impact. We could drastically reduce runtime and overall considered features to replicate slightly better results in predicted correct answers per task. This successfully answers our first research question defined in this section. It is also important to note that Frankenstein blindly traverses all the question features and uses the same features for all tasks. However, this experiment concludes that for each task, the local optimisation is impacted by different features. For example, boolean answer type of the question as a feature does not have any impact on NED task whereas for QB task, this feature has a significant impact (cf.\ \figurename~\ref{fig:old-gini}). The effectiveness and efficiency of feature selection per task for Frankenstein 2.0 settings compared to Frankenstein is illustrated in \figurename~\ref{fig:old-gini11}. Selecting the most significant features not only reduces execution time but also improves the number of predicted top N questions per task. 

\subsection{Experiment 2: Impact of Integrating New Components on Local Optimisation} \label{sec:experiment2}
\begin{figure}
    \includegraphics[scale=0.70]{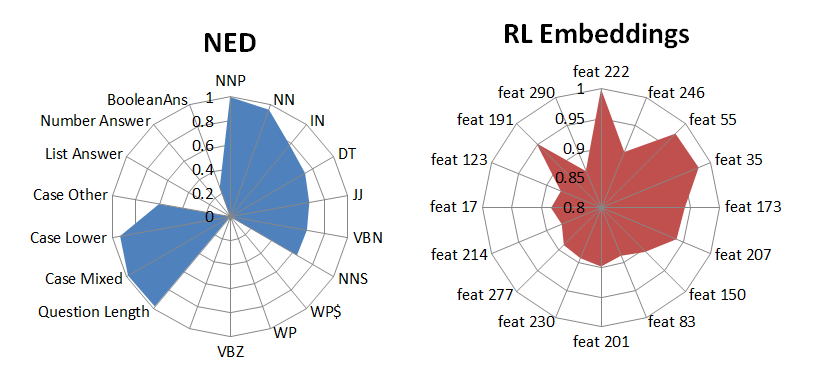}
    \caption{\textbf{Significant features per QA task after integrating new components for NED and RL (cf.\ \autoref{sec:experiment2})}. Gini importance is used to calculate the feature significance (higher is better). Each task is impacted by different features, e.g.\ the NED task is impacted most by NLP features (CF2), while for the RL task, Fasttext embeddings have most impact (CF3).}
    \label{fig:new-gini}
\end{figure}

In this experiment, we address the second research question \emph{R2: What is the impact of integrating high performing components on local performance (i.e.\ task level)?} Please note that in this experiment, the prediction model among Frankenstein 2.0 and Frankenstein remains the same.
During the last year, multiple open source QA components were released such as EARL~\cite{DBLP:conf/semweb/DubeyBCL18} and Falcon~\cite{falcon} which accomplish both NED and RL tasks and the second version of Ambiverse\footnote{\url{https://developer.ambiverse.com/docs}} for NED task. We benchmarked these tools on 3,253 questions of LC-QuAD and report that EARL, Falcon, Ambiverse NED components have F-score 0.54, 0.73, 0.65 respectively; for the RL task, EARL and Falcon have F-score 0.27 and 0.56 respectively. Recently component SQG~\cite{DBLP:conf/esws/ZafarNL18} for the QB task has also been released. Please note that SQG uses 85\% of LC-QuAD questions for training, which hinders us to perform 10-fold validation over 3,000 questions of LC-QuAD and, therefore, we do not include this component in the experiment.
Shortly after, these components were integrated into Frankenstein which contained 21 NED, seven RL, two CL and two QB components in total. To the best of our knowledge, there is no other related open source component which has not been integrated.
We run experimental studies on the impact of adding new components on the local performance (task-level benchmarking for RL and NED tasks because only these tasks integrated new components). 
Furthermore, since the local optimisation in any CQA framework is influenced by the performance of the classifiers as well as the performance of components, we run feature engineering once more after adding the new components. Our goal is to figure out the optimum set of features per task because \textit{the newly added components might change the prominent features}.
To achieve this, we experiment with the impact of each feature configuration (CF1--CF6) on the prediction model of the updated list of components. We empirically selected top N (N=15) impacting features per task as illustrated in \figurename~\ref{fig:new-gini}.
This experiment revealed that adding new components led to reordering the set of the impacting feature configuration. 
For instance, for the NED task, question length and character cases and for the RL task, embeddings emerged to be the most impacting features comparing to the previous experiment (\figurename~\ref{fig:old-gini}). 

The next experiment is about evaluating the impact of new components on Frankenstein 2.0. We use the following settings:

\begin{figure}
    \centering
    \includegraphics[scale=0.60]{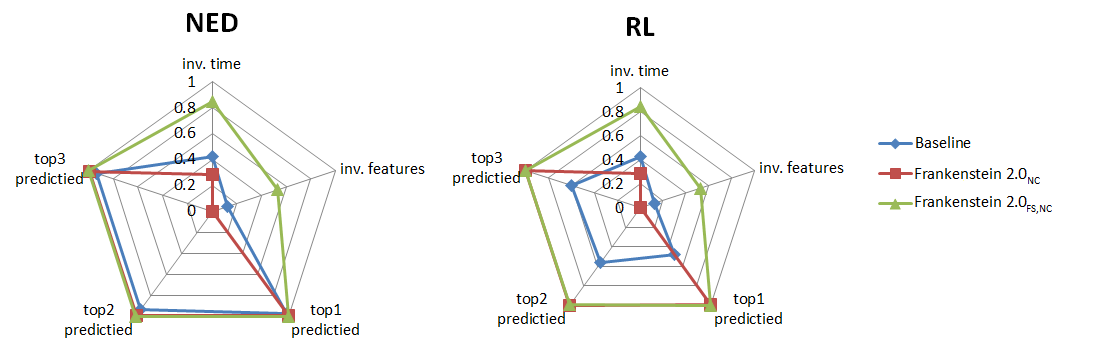}
    \caption{\textbf{Impact of Adding New Components to Frankenstein}. A significant jump in the overall performance with respect to the baseline is observed when three NED and two RL components are added. Improvement is measured in terms of three metrics: 1) normalised number of top 1, 2 and 3 predicted questions; 2) inverse of normalised number of selected features; and 3) and inverse of normalised execution time (training and predicting using ML model). For all the metrics: higher is better.}
    \label{fig:experiment-2-comparision}
\end{figure}
\begin{enumerate}
     \item Baseline: Frankenstein baseline from~\cite{DBLP:conf/www/SinghRBSLUVKP0V18}.
    \item Frankenstein 2.0$_{NC}$: integrating new components and applying the similar setting of the baseline (same features and prediction model). 
    \item Frankenstein 2.0$_{FS,NC}$: integrating new components and applying optimum features (i.e.\ CF2 for NED and CF3 for RL).
\end{enumerate}

Table \ref{tab:newcomponent} presents the performance (using "\#answerable" and "predicted top N component" metrics) of Frankenstein 2.0.
This experiment shows that Frankenstein 2.0$_{FS,NC}$ setting outperforms the others. 
This enhancement for RL task is bolder (i.e.\ more than 96 questions).
The improvement for Frankenstein 2.0$_{NC}$ setting is slighter. \\
\textbf{Experiment Conclusion}: In this experiment, we studied the impact of new QA components on the framework performance. We report that the addition of components also results in the reordering of features that impact the task level performance (i.e.\ local optimisation) in terms of "predicted top N component". As observed in \figurename~\ref{fig:new-gini}, adding a new RL component completely changed the features that impact the task level performance and result in the reordering of feature impact for NED tasks. These findings reveal that for any automatic CQA, dynamic (on-the-fly) feature selection per task is necessary rather than keeping a fixed feature set per task.

\figurename~\ref{fig:experiment-2-comparision} summarises our overall findings for this experiment. We conclude that Frankenstein 2.0$_{FS,NC}$ setting significantly improves the task level performance (i.e.\ local optimisation) in terms of "predicted top N component". In addition, this setting requires lower execution time (in training and testing) and fewer number of features.


\begin{table}[tb] 
	\centering
	\caption{Evaluation of the Impact of Integrating New Components in Frankenstein 2.0. Adding new components for NED and RL tasks leads to higher performance.}
	\resizebox{\columnwidth}{!}{%
	    \begin{tabular}{ l c c c c c c}
	        \toprule
		    \textbf{QA} & \multicolumn{1}{l}{\textbf{\#totalfold}} & \multicolumn{1}{l}{\textbf{\#answer-}} & \multicolumn{3}{c}{\textbf{\#Predicted}} & \multicolumn{1}{l}{\textbf{Setting}} \\
		    \cline{4-6}
		    \textbf{Task} & \multicolumn{1}{l}{\textbf{}} & \multicolumn{1}{l}{\textbf{able}} & \multicolumn{1}{l}{\textbf{Top1}} & \multicolumn{1}{l}{\textbf{Top2}} & \multicolumn{1}{l}{\textbf{Top3}} & \multicolumn{1}{l}{\textbf{}} \\ 
		    \midrule
		    NED & 324.3 & 294.2 & 245.2 & 270.9 & 284.3 & \multicolumn{1}{c}{Baseline}  \\
		    NED & 324.3 & 309.1 & \underline{250.5}& \underline{286.8} & \underline{298.1} & \multicolumn{1}{c}{Frankenstein 2.0$_{NC}$}  \\
		    NED & 324.3 & 309.1 & \underline{252.6} & \underline{288.1} & \underline{299.5} &  \multicolumn{1}{c}{\textbf{Frankenstein 2.0$_{FS,NC}$}} \\
		    RL & 324.3 & 153.1 & 90.3 & 118.9 & 134.4 & \multicolumn{1}{c}{Baseline}  \\
		    RL & 324.3 & 231.7 &\underline{186.5}& 213.1 & 224.9 & \multicolumn{1}{c}{Frankenstein 2.0$_{NC}$}  \\
		    RL & 324.3 & 231.7 & \underline{187.7} & 211.1 & \underline{225.7} & \multicolumn{1}{c}{\textbf{Frankenstein 2.0$_{FS,NC}$}} \\
		     \bottomrule
		      	\end{tabular}
		      	}
	\label{tab:newcomponent}   
\end{table}


\subsection{Experiment 3: Impact of Learning Models on Local Optimisation} \label{sec:ml}
\begin{table}[tb] 
	\centering
	\caption{Evaluation of the Impact of Supervised Learning Models. For NED and RL, Random Forest performs slightly better than Logistic Regression.}
	\resizebox{\columnwidth}{!}{%
	    \begin{tabular}{ l c c c c c c}
	        \toprule
		    \textbf{QA} & \multicolumn{1}{l}{\textbf{\#totalfold}} & \multicolumn{1}{l}{\textbf{\#answer-}} & \multicolumn{3}{c}{\textbf{\#Predicted}} & \multicolumn{1}{l}{\textbf{Setting}} \\
		    \cline{4-6}
		    \textbf{Task} & \multicolumn{1}{l}{\textbf{}} & \multicolumn{1}{l}{\textbf{able}} & \multicolumn{1}{l}{\textbf{Top1}} & \multicolumn{1}{l}{\textbf{Top2}} & \multicolumn{1}{l}{\textbf{Top3}} & \multicolumn{1}{l}{\textbf{}} \\ 
		    \midrule
		    NED & 324.3 & 294.2 & 245.2 & 270.9 & 284.3 & \multicolumn{1}{c}{Logistic Regression}  \\
		    NED & 324.3 & 309.1 & \underline{249.5}& \underline{270.5} & \underline{283.6} & \multicolumn{1}{c}{\textbf{Random Forest}}  \\
		    RL & 324.3 & 153.1 & 90.3 & 118.9 & 134.4 & \multicolumn{1}{c}{Logistic Regression}  \\
		    RL & 324.3 & 231.7 &\underline{91}& 116.6 & 134.3 & \multicolumn{1}{c}{\textbf{Random Forest}}  \\
		    CL & 324.3 & 76 & 68.1 & 76 & \multicolumn{1}{c}{–}& \multicolumn{1}{c}{Logistic Regression}  \\
		    CL & 324.3 & 76 & \underline{68.8} & 76 & \multicolumn{1}{c}{–}& \multicolumn{1}{c}{\textbf{Random Forest}}  \\
		    QB & 324.3 & 175.4 & \underline{162.7} & 175.4 & \multicolumn{1}{c}{–}& \multicolumn{1}{c}{\textbf{Logistic Regression}} \\
		    QB & 324.3 & 175.4 & 159.5 & 175.4 & \multicolumn{1}{c}{–}& \multicolumn{1}{c}{Random Forest} \\
		     \bottomrule
		      	\end{tabular}
		      	}
	\label{tab:MLmodels}   
\end{table}

In this experiment, we address the third research question \emph{R3: What is the impact of employing a well-fitted learning model on local performance?} To do that, we ran various supervised learning models to find out the best performing one. These models include: Logistic Regression, Gradient Boosting, XGBoosting, PCA + Logistic Regression, SVM, Naive Base, LDA, Adaboost and feed-forward neural network.  Each Model is tuned to its optimum performance. Table \ref{tab:MLmodels} summarises the results for the top-2 models for each task. The rest of the results can be found in our Github project.

For demonstrating the impact of ML models compared to the baseline, we use following settings:
\begin{enumerate}
     \item Baseline setting of~\cite{DBLP:conf/www/SinghRBSLUVKP0V18}.
     \item Frankenstein 2.0$_{ML}$: applying Random Forest algorithm for NED and RL, Logistic Regression for CL and QB along with the settings of Baseline (keeping features and number of components constant). 
    \end{enumerate}
Table \ref{tab:newmodel} shows that Frankenstein 2.0$_{ML}$ setting performs slightly better than the Baseline setting for the NED and RL tasks whereas it exposes equivalent performance for the CL and QB tasks. \\
\textbf{Experiment Conclusion:} In this experiment, we followed a benchmarking approach for supervised learning models to choose the best model per task. Frankenstein uses logistic regression for all tasks. We have observed that for NED and RL, Random Forest performs slightly better than Logistic Regression. However, performance improvement is not significant for any task. We conclude that for an automatic CQA, only the choice of the ML model will not have a huge impact. In the previous experiments, we observed that feature engineering and the addition of newer components have impacted the performance. These findings motivate us to combine our individual strategies together to study the overall impact on the Frankenstein 2.0 performance. We detail our findings in the next experiment.

\subsection{Experiment 4: Impact of all Strategies of Frankenstein 2.0 on the Overall Local Performance} \label{sec:ml2}
In this experiment, we address the fourth research question \emph{R4: What is the impact of employing Frankenstein 2.0 (local optimiser consists of feature engineering, new components and choice of ML models) on the overall local performance?}
To answer that, we provide an empirical study using the three following settings: \begin{enumerate}
     \item Baseline setting of~\cite{DBLP:conf/www/SinghRBSLUVKP0V18}.
    \item Frankenstein 2.0$_{FS,NC,ML}$: applying Random Forest algorithm for NED and RL and Logistic Regression for CL and QB along with the settings of Frankenstein 2.0$_{FS,NC}$ (i.e.\ feature engineering and addition of new components). 
    \item Frankenstein 2.0: Frankenstein 2.0$_{FS,NC,ML}$ setting with a reduced number of components for NED and RL tasks.
    
    \end{enumerate}
    
\begin{table}[tb] 
	\centering
	\caption{Overall Improvement at Task Level from Baseline after Component Addition. Frankenstein 2.0 achieves similar performance compared to Frankenstein 2.0$_{FS,NC,ML}$ even when the number of components is reduced.}
	\resizebox{\columnwidth}{!}{%
	    \begin{tabular}{ l c c c c c c}
	        \toprule
		    \textbf{QA} & \multicolumn{1}{l}{\textbf{\#totalquestions}} & \multicolumn{1}{l}{\textbf{\#answer-}} & \multicolumn{3}{c}{\textbf{\#Predicted}} & \multicolumn{1}{l}{\textbf{Setting}} \\
		    \cline{4-6}
		    \textbf{Task} & \multicolumn{1}{l}{\textbf{}} & \multicolumn{1}{l}{\textbf{able}} & \multicolumn{1}{l}{\textbf{Top1}} & \multicolumn{1}{l}{\textbf{Top2}} & \multicolumn{1}{l}{\textbf{Top3}} & \multicolumn{1}{l}{\textbf{}} \\ 
		    \midrule
		    NED & 324.3 & 294.2 & 245.2 & 270.9 & 284.3 & \multicolumn{1}{c}{Baseline}  \\
		    NED & 324.3 & 294.2 & 249.5 & 270.5 & 283.6 & \multicolumn{1}{c}{Frankenstein 2.0$_{ML}$}  \\
		    NED & 324.3 & 309.1 & \underline{257.5}& \underline{289.8} & \underline{301.3} & \multicolumn{1}{c}{Frankenstein 2.0$_{FS,NC,ML}$}  \\
		    NED & 324.3 & 309.1 & \underline{257.3} & \underline{289.5} & \underline{301.2} &  \multicolumn{1}{c}{\textbf{Frankenstein 2.0}} \\
		    RL & 324.3 & 153.1 & 90.3 & 118.9 & 134.4 & \multicolumn{1}{c}{Baseline}  \\
		     RL & 324.3 & 153.1 & 91.6 & 119.9 & 134.9 & \multicolumn{1}{c}{Frankenstein 2.0$_{ML}$}  \\
		    RL & 324.3 & 231.7 &\underline{188}& \underline{214} & \underline{224.6} & \multicolumn{1}{c}{Frankenstein 2.0$_{FS,NC,ML}$}  \\
		    RL & 324.3 & 309.1 & \underline{188} & \underline{213.9} & \underline{226.1} & \multicolumn{1}{c}{\textbf{Frankenstein 2.0}} \\
		    CL & 324.3 & 76 & 68.1 & 76 & \multicolumn{1}{c}{–}& \multicolumn{1}{c}{Baseline}  \\
		    CL & 324.3 & 76 & \underline{68.9} & 76 & \multicolumn{1}{c}{–}& \multicolumn{1}{c}{\textbf{Frankenstein 2.0$_{FS,NC,ML}$}}  \\
		    QB & 324.3 & 175.4 & 162.7 & 175.4 & \multicolumn{1}{c}{–}& \multicolumn{1}{c}{Baseline} \\
		    QB & 324.3 & 175.4 & 162.6 & 175.4 & \multicolumn{1}{c}{–}& \multicolumn{1}{c}{\textbf{Frankenstein 2.0$_{FS,NC,ML}$}} \\
		     \bottomrule
		      \end{tabular}}
	\label{tab:newmodel}    
\end{table}

\begin{figure*}[t]
    \centering
    \subfloat{
        \includegraphics[width=\textwidth]{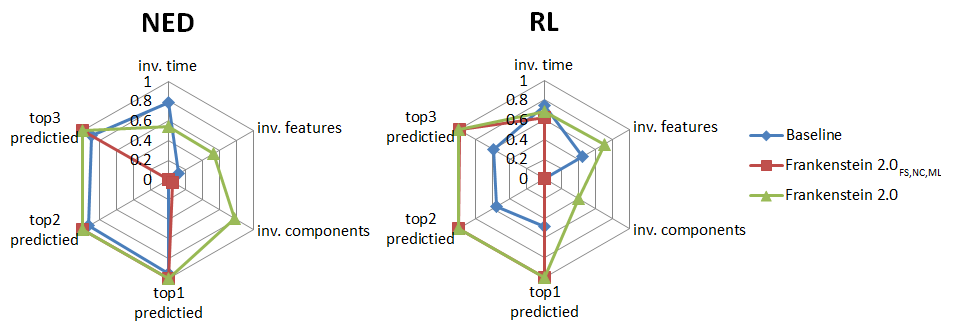}
        }
  \hfill
  \subfloat{
    \includegraphics[width=\textwidth]{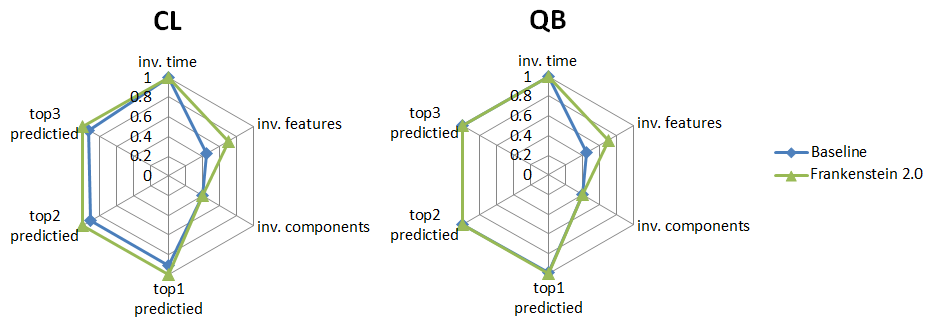}
    }
  \caption{
	\textbf{Overall Improvement}. There are 18 NED, seven RL, two CL and two QB components. Frankenstein 2.0 best setting significantly outperforms Frankenstein in terms of efficiency (shown as normalised number of top N predicted questions) and effectiveness (normalised inverse execution time and normalised inverse number of selected features. All the metrics are: higher is better.
	}
	\label{fig:experiment-3-comparision}
\end{figure*}

The Frankenstein 2.0$_{FS,NC,ML}$ setting considerably outperforms the Baseline and Frankenstein 2.0$_{ML}$ settings. For instance, for NED it has an improvement of 12 questions and for RL it reaches 98 questions which is an impressive achievement. We then reduced the number of components in Frankenstein 2.0 setting. The components are reduced on decreasing F-score of components over the LC-QuAD dataset. The rationale behind that is that a reduced number of components might result in a similar performance while reducing the overhead. Thus, Frankenstein 2.0 reduces NED components to top five and RL components to the top three. We observed this setting empirically. The comparison (cf.\ Table \ref{tab:newmodel}) shows that Frankenstein 2.0 achieves similar performance (with respect to top-N correctly predicted questions) compared to Frankenstein 2.0$_{FS,NC,ML}$ even when the number of components is reduced. This behaviour appears due to the fact that the newly integrated components have comparable behaviour with the remaining components with regard to the questions they can address(not all the components are complementary, similar behaviour was observed in~\cite{DBLP:journals/corr/abs-1809-10044}). This was not the case for CL and QB tasks for which the components are complementary.\\
\textbf{Experiment Conclusion}: This experiment summarises our three-fold extension of Frankenstein as Frankenstein 2.0. We included a new feature selection module that reduces the unnecessary features per task. We added new components and found that the newly added components for NED and RL are not complementary and we benchmarked supervised ML models. It is important to note that traditionally CQA frameworks aim to integrate a number of components (OKBQA and Frankenstein have 24 and 28 components respectively). We observe that adding more components not necessarily results in higher performance. Adding high performing component coupled with feature engineering can significantly improve the performance of an automatic CQA. Hence, for designing automatic CQA frameworks, dynamic (on-the-fly) approach is needed not for just selecting the components, but also for selecting the most prominent features per question. With the addition of new components in the CQA framework, an automatic CQA should also reorder the features set per task. Also, it is observed in our experiments that performance of supervised learning methods do change per task. Hence, for any CQA framework, dynamic composition of QA pipelines is needed at three levels: 1) selection of most prominent question features, 2) selection of best components per task for each input question and 3) selection of best supervised learning model per task. 



\section{Discussion and Conclusion Remarks}\label{sec:Conclusions}
In this paper, we proposed Frankenstein 2.0 to solve the \textbf{local optimisation problem} (task-level performance) of an automatic CQA framework and extended Frankenstein in three directions. In order to reach the highest performance, Frankenstein 2.0 is proposed after running careful empirical studies on feature engineering and machine learning models and determining the impact of integrating new components. After exhaustive evaluation, we finally compared Frankenstein 2.0 against the Baseline with respect to three metrics: 1) execution time of training and testing the learning model, 2) number of components used and 3) and number of questions answered (\figurename~\ref{fig:experiment-3-comparision}).
Applying Frankenstein 2.0 on the state-of-the-art CQA framework leads to significant improvements specifically for the RL task which is one of the major bottlenecks inhibiting CQA frameworks~\cite{DBLP:conf/www/SinghRBSLUVKP0V18}. Although we ran a detailed study on finding the best supervised learning approach, the performance improvement was not significant. We conclude that feature selection and component addition coupled with the best supervised learning model results in significant improvement of task-level performance. 

It is important to note that addition of new components does not necessarily improve the task-level performance. The improvement depends on two factors: 1) adding new complementary high performing components and  2) adding a local optimiser, such as Frankenstein 2.0, which takes care of best setting per task and chooses the number of components wisely.  Throughout our experiments, we learned that a local optimiser requires a dynamic feature engineering and choice of the learning model. Therefore, for any upcoming component, the local optimisation techniques integrated into a CQA framework can learn the most impactful features as well as the learning model. Furthermore, for any ecosystem of CQA, it can find the optimum number of components, thus reducing the overhead on the pipeline without any performance loss. We admit, CQA frameworks need a lot of effort in the research community to be able to compete with end to end performance of monolithic QA systems. This article successfully attempts to solve an important part of the overall problem. We plan to extend our work by proposing global optimisation strategies and evaluate the QA pipelines against state-of-the-art QA systems.

%

\balance


\section*{References}
\bibliographystyle{ACM-Reference-Format}
\bibliography{references}

\end{document}